# Kinect Calibration and Data Optimization For Anthropometric Parameters

M.S.GOKMEN [1], M.AKBABA [2], O.FINDIK [3]

[1] Karabuk University, Karabuk /Turkey, selmangokmen@karabuk.edu.tr
[2] Karabuk University, Karabuk /Turkey, mehmetakbaba@karabuk.edu.tr
[3] Karabuk University, Karabuk /Turkey, oguzfindik@karabuk.edu.tr

*Abstract* - **Recently, through development of several 3d vision systems, widely used in various applications, medical and biometric fields. Microsoft kinect sensor have been most of used camera among 3d vision systems. Microsoft kinect sensor can obtain depth images of a scene and 3d coordinates of human joints. Thus, anthropometric features can extractable easily. Anthropometric feature and 3d joint coordinate raw datas which captured from kinect sensor is unstable. The strongest reason for this, datas vary by distance between joints of individual and location of kinect sensor. Consequently, usage of this datas without kinect calibration and data optimization does not result in sufficient and healthy. In this study, proposed a novel method to calibrating kinect sensor and optimizing skeleton features. Results indicate that the proposed method is quite effective and worthy of further study in more general scenarios.**

*Keywords* - **Kinect Sensor, Anthropometry, Joint, Optimization, 3d vision systems**

## I. INTRODUCTION

In the last decade, by development of 3d vision systems, have been made available for scientific applications. Most important feature of 3d vision systems (RGBD cameras) is that capturing distance close to real world distances approximately [1]. This feature made Kinect sensor available for medical and biometric applications [2]. In 2010, Microsoft company released a new RGBD camera that called Kinect v1 sensor. After, Microsoft company released Kinect Sensor v2 that contains Time of Flight (ToF) technology in 2014. Cause of Kinect sensor affordable among other sensors, Kinect sensor have been made available that for application development [3]. Another major factor that made Kinect sensor popular, providing for 3d coordinate datas as real-like [4].

Studies made with Kinect sensor, substantially focused on gait recognition, gesture recognition and biometric analysis. Used data sets in this studies that are usually unexceptional and standart [5]. Cause for this, reducing inaccuracy arising out of by Time of Flight technology. Therefore, for improving data quality that obtained from Kinect sensor, calibration and data optimization is essential.

In this study, provided calibration of Kinect sensor and optimization of datas that obtained by Kinect sensor. In the first stage of calibration, measured distance between Kinect sensor and ground. Therefore, aimed to establish height of joint points. The dataset used for Kinect calibration is determined as vertical and horizontal to Kinect sensor on XZ space. Calibration data were collected as 10 times vertically to XZ space and 10 times horizontally to XZ space.

## II. RELATED WORKS

Consistency of anthropometric parameters is very important for improvement of studies in gait recognition, gesture recognition and biometric fields. The most common used parameters are temporal parameters, spatial parameters and kinematic parameters in gait and gesture recognition [6]. Spatial and temporal parameters are the intuitive gait features including step length, speed, gait cycle, average stride length, and so on. Kinematic parameters are usually characterized by the joint angles between body segments and their relationships to the events of the gait cycle [6]. Anthropometric parameters are individual features including like bone lengths [7].

Accuracy of spatial and temporal parameters is related to accuracy of kinematic parameters. In situation that joint points coordinate data are incompatible, can not expected being result healthy of length features like stride lengths. In some studies, for preserving of kinematic parameters accuracy, individuals was asked to walking on straight path [8]. Thus, the data collected as least affected from noise of unexpected movements. Movement of contrary to expectations, can effect to consistence of parameters. Therefore, Kinect calibration is necessary for accuracy of parameters that obtained from individuals that in anywhere in Kinect sight [9].

Measurements between joint points are utilized in gait recognition or gesture recognition fields [10]. In related studies, relative distance features (RDF) and vertical distance features (VDF) are particularly used [4][5]. Obtaining coordinate data of joint points as accurate is very important for healthy resulting of study. Thus, importance of Kinect sensor calibration and optimization of data obtained from Kinect sensor that is essential.

Anthropometric parameters are acquired from in consequence of processing of spatial, temporal and kinematic parameters. Anthropometric parameters are fundamental values in performed applications that worked on biometric and identification applications [7]. For accurate of obtained anthropometric parameters from Kinect sensor, consistency of spatial and temporal parameters are crucial. The improperly parameters that obtained from Kinect sensor, results wrong classification and wrong bone lengths calculation. In studies that used anthropometric parameters, does not performed any application for demonstration of anthropometric parameters accuracy [12]. Reason for this, generation of unstable parameters from non calibrated Kinect sensor. Therefore, unstable and low accuracy parameters must be optimized. Importance of clear and stable joint points that obtained from Kinect sensor, is bone lengths are uniquely by individual. Hence, importance of anthropometric parameters are seen obviously that used in identification applications.

## III. METHODOLOGY

Kinect sensor v2 that is used in this study, developed for X-box 360 video console and it connected to pc by usb converter and works with SDK of version 2.0. The Kinect sensor equipped with RGB camera, light emitter and infrared-sensitive depth sensor. Parameters that obtained from Kinect sensor, processes by software library that known NUI API (Natura User Interface). This API made available that 3d data of individuals. The data obtained from Kinect sensor by NUI API that is 25 joint points which occurs coordinates of X,Y and Z that have totally 75 piece of data [11]. Joint points represent as joint points and real world coordinates that belongs to individual. Through usage of SDK, can be achieve data with 30 frame per second. With every captured frame, the data is obtained that contains joint points stored in an array.

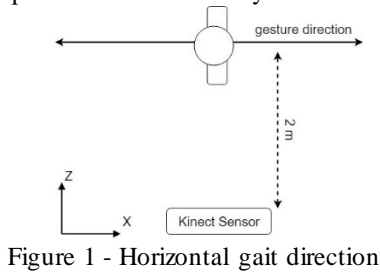

Figure 1 - Horizontal gait direction

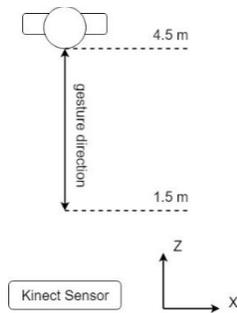

Figure 2: Vertical gait direction

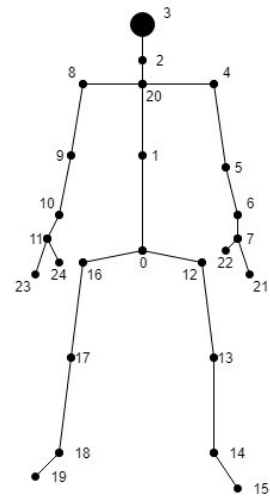

Figure 3 - Joint Points

| Joint | Value |
|-------|-------|
| Base of the spine | 0 |
| Middle of the spine | 1 |
| Neck | 2 |
| Head | 3 |
| Left shoulder | 4 |
| Left elbow | 5 |
| Left wrist | 6 |
| Left hand | 7 |
| Right shoulder | 8 |
| Right elbow | 9 |
| Right wrist | 10 |
| Right hand | 11 |
| Left hip | 12 |
| Left knee | 13 |
| Left ankle | 14 |
| Left foot | 15 |
| Right hip | 16 |
| Right knee | 17 |
| Right ankle | 18 |
| Right foot | 19 |
| Spine at the shoulder | 20 |
| Tip of the left hand | 21 |
| Left thumb | 22 |
| Tip of the right hand | 23 |
| Right thumb | 24 |

Table 1: Joint Point Index

First, data were collected vertically to Kinect sensor on XZ coordinate space as shown on fig. 2. When collected datas observed, determined joint point coordinates have irregularity on it. The reason for this is angle of inclination between Kinect sensor and X coordinate axis. By variance of inclination angle, Kinect sensor's coordinate axis change according to inclination angle. Height of Kinect sensor from ground represented as $h_k$ on fig. 5.

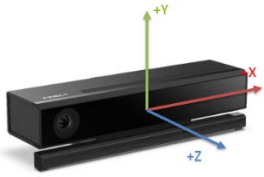

Figure 4 : Kinect coordinate axis

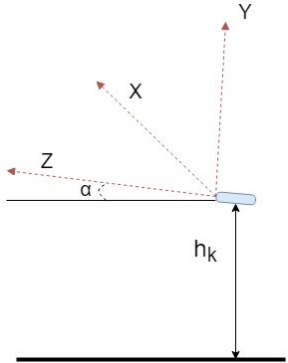

Figure 5: Kinect with inclination angle

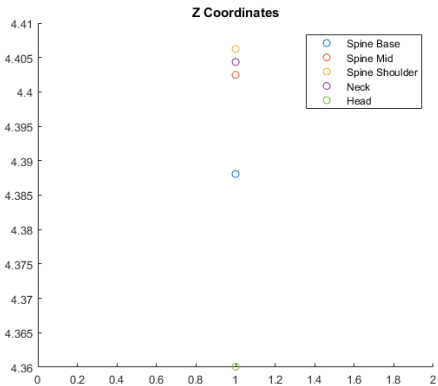

Figure 6: Z coordinates of Joints

In fig. 6, shown Z coordinate values of some specified joint points gathered from individual which standing. Inconsistency that occurs by inclination angle is noticeable in fig. 6.

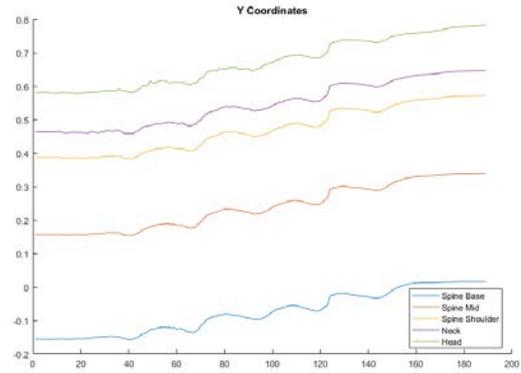

Figure 7: Y coordinate variation

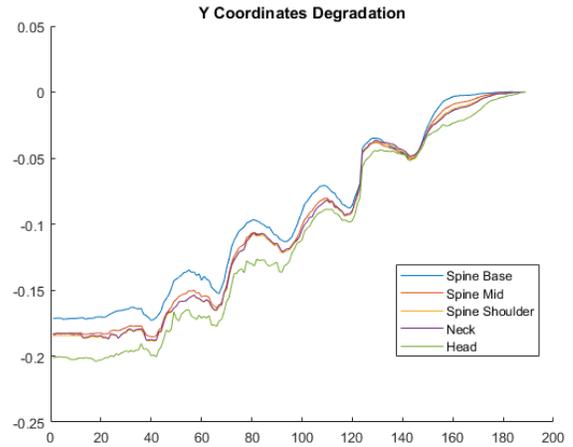

Figure 8: Difference between Y coordinate of last frame and Y coordinate of each frame with raw data

As shown on fig. 7, Y coordinate variations of specified joint points are not consistent caused by inclination angle. To solve this problem, inclination angle must be calculated. In fig. 8, clearly visible variety of Y coordinate values between last frame and each frame.

In our method, while resolving inclination angle, used two spine joint points that are base of the spine and middle of the spine. If inclination angle represents as α, according to every performed gait for calibration, detected least inconsistency on these joint points. Inclination angle was represented as $\propto_m^b$ that for every performed calibration gait. In function that shown at above, $b$ variance is represented as calibration gaits. Inclination angle (α) is calculated for each frame in calibration and represented as $\propto_i$ . $n$ variable represents frame number for calibration gaits. $m$ variable shown as 'mean'.

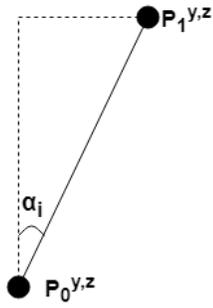

$$\propto_m^b = \frac{\sum_{i=1}^n \propto_i}{n} \qquad \textbf{Eq.1}$$

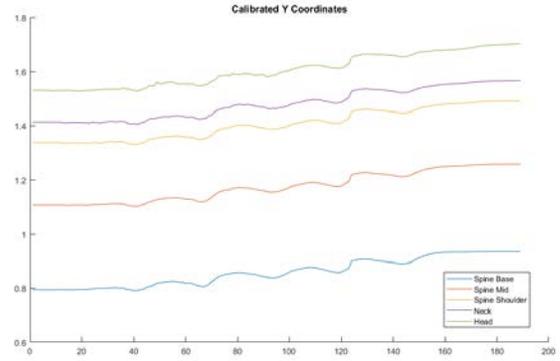

Figure 10: Calibrated Y coordinates

Inclination angle that intended to use for calibration ($\propto_g$), represented as geometric mean of each obtained arithmetic mean of inclination angle from calibration gaits.

$$\propto_g = (\prod_{i=1}^n \propto_m^i)^{1/n} \qquad \textbf{Eq. 2}$$

By calculation of mean inclination angle ($\propto_g$), could improve inconsistencies on Y and Z coordinate axis. Y and Z coordinate datas that belong to each joint points is represented as in equation at below.

$$J_i^p(i_0:i_n) = <P_i^x, P_i^y, P_i^z> \qquad \textbf{Eq. 3}$$

In equation at above, $i$ variable represents as index number of each joint point at table 1. In equation 4, new joint point coordinates that calibrated according to $\propto_g$ are represented as in equation at below.

$$J_{i,c}^p(i_0:i_n) = <P_{i,c}^x, P_{i,c}^y, P_{i,c}^z> \qquad \textbf{Eq. 4}$$

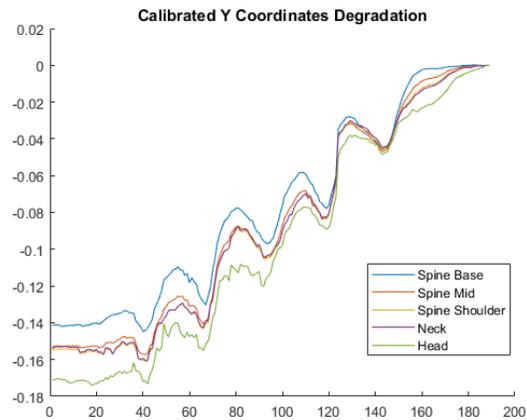

Figure 11: Difference between Y coordinate of last frame and Y coordinate of each frame with calibrated data

In Figure 10, joint points that calibrated has less inconsistency. Whereas, the joint points which located in higher Y coordinates has more inconsistency than located in lower Y coordinates. Cause of this perspective. Depending on perspective, observed that variance of joint point coordinates on individuals which getting closer to Kinect sensor. If assumed that equation $P_{i,c}(i_0:i_n) = <C_i^x, C_i^y, C_i^z>$, $i$ represents index number of each joint point.

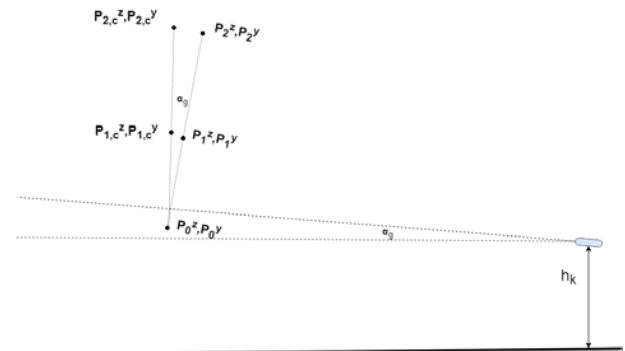

Figure 9: Calibration of Kinect sensor

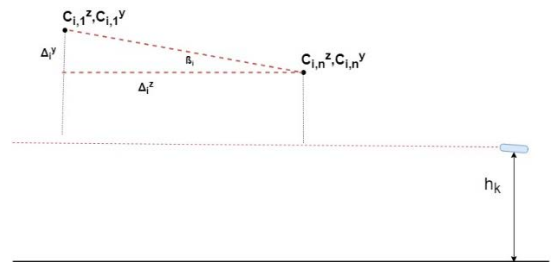

Figure 12: Perspective degree

$$P_{i,c}^z = P_i^y * \sin(\propto_g) + P_i^z \qquad \textbf{Eq. 5}$$

$$P_{i,c}^y = P_{i,c}^z * \sin(\propto_g) + P_i^y + h_k \qquad \textbf{Eq. 6}$$

In figure 12, $C_{i,1}^y$, $C_{i,1}^z$ and $C_{i,n}^y$, $C_{i,n}^z$ variables represent that Y and Z coordinate values of specified joint points at first and last frame. Difference between coordinate values at first and last frame has shown as $\Delta_i^y$ and $\Delta_i^z$ in figure 12. After implementation of this variable, perspective degree ($ß_i$) is calculated shown as equation 7. After calculation of perspective degree ($ß_i$) for each calibration gait, mean perspective degree is represented as $ß_{i,m}$ in equation 8. In equation 8, $b$ represents each calibration gait and $i$ represents indexes of each joint points. For purpose of calculation of $ß_i$ value, joint points were elected according to Y coordinate values that are going from top to bottom in Y coordinate axis.

$$ß_i = tan^{-1}\left(\frac{C_{i,1}^y - C_{i,n}^y}{C_{i,1}^z - c_{i,n}^z}\right) \qquad \textit{Eq. 7}$$

$$ß_{i,m} = \frac{\sum_{n=1}^{b} ß_{i,n}}{b} \qquad \textit{Eq. 8}$$

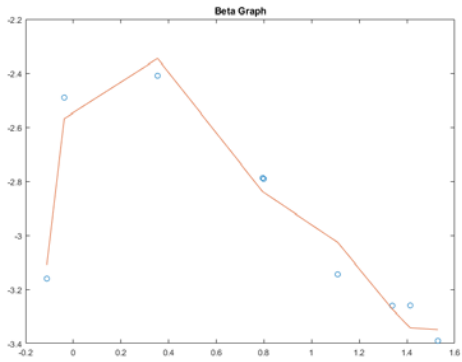

Figure 13: Beta Graph

Variance of perspective degrees according to Y coordinate values of calibrated joint points has shown in figure 13. $P_ß$ polynomial function was obtained by implementation of curve fitting from over $ß_i$ points that shown in figure 13. Through, can be found perspective degree related to height of Y coordinate axis.

$$ß_{i,n}^y = P_ß(C_{i,n}^y)$$
$$C_{i,n}^y = C_{i,n}^y + C_{i,n}^z * tan(ß_{i,n}^y) \qquad \textit{Eq. 9}$$

In equation 9, new Y coordinate values of calibrated joint points was calculated with $P_ß$ polynomial equation. $n$ represents frame number in equation and $i$ represents index number of joint points.

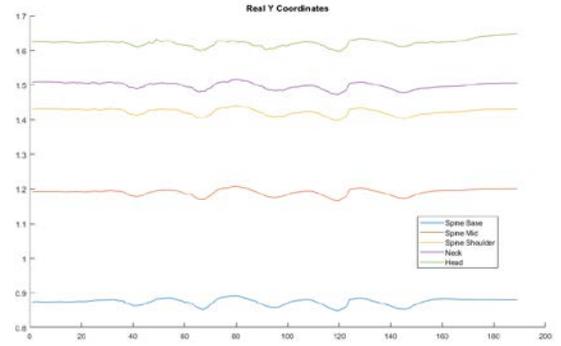

Figure 14: Y coordinate values of optimized data

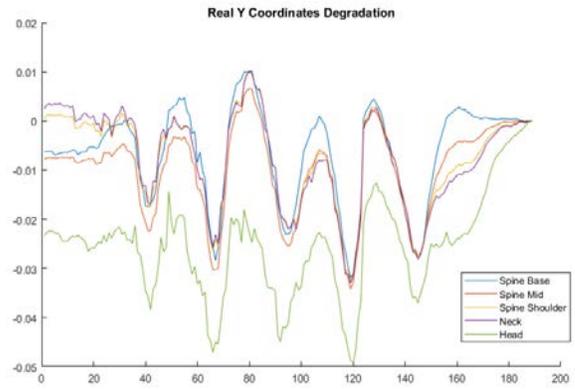

Figure 15: Difference between Y coordinate of last frame and Y coordinate of each frame with optimized data

Experimental results show that, our proposed method optimized data successfully. As shown in figure 14, Y coordinate value of joint points are stable. Reason for big majority of fluctuations is only from gait. Maximum difference variety is clearly seen obviously as 5 cm in figure 15.

## IV. CONCLUSIONS

As a result of study, Kinect calibration and joint coordinate data optimization successfully achieved. In study, achieved data optimization of joint point coordinate values over no filter applied on it. This study proposes to contribute to studies will made with Kinect Sensor.

This work supported by Research Fund of the Karabuk University. Project Number: KBÜBAP-17-YL-250